# Correction of "CLOUD REMOVAL BY FUSING MULTI-SOURCE AND MULTI-TEMPORAL IMAGES"

*Chengyue Zhang[1], Zhiwei Li[1], Qing Cheng[2], Xinghua Li[3], Huanfeng Shen[1]*

[1]School of Resource and Environmental Sciences, Wuhan University, China
[2]School of Urban Design of Wuhan University, Wuhan University, China
[3]School of Remote Sensing and Information Engineering, Wuhan University, China

## ABSTRACT

Remote sensing images often suffer from cloud cover. Cloud removal is required in many applications of remote sensing images. Multitemporal-based methods are popular and effective to cope with thick clouds. This paper contributes to a summarization and experimental comparison of the existing multitemporal-based methods. Furthermore, we propose a spatiotemporal-fusion with poisson-adjustment method to fuse multi-sensor and multi-temporal images for cloud removal. The experimental results show that the proposed method has potential to address the problem of accuracy reduction of cloud removal in multi-temporal images with significant changes.

## 1. INTRODUCTION

Cloud cover is generally present in remote sensing images, which limits the potential of the images for ground information extraction. Therefore, removing the clouds and recovering the ground information for the cloud-contaminated images is often necessary in many applications. Much research effort has been devoted to the task of cloud removal for remote sensing images.

Cloud removal is essentially an information reconstruction process, and the reconstruction approaches can be grouped into three different categories according to the different sources of the complementary information used [1]. One category is the spatial interpolation based approaches which use the remaining parts in the image to predict the cloud-contaminated regions, without the aid of other complementary data. The reconstruction results of this category of methods are often visually plausible but with low accuracy, difficult to meet the application requirements [2]. The second category is multispectral-based approaches which restore the cloud-contaminated image by using complementary information of the multispectral bands [3]. However, this category of methods tends to remove thin clouds but have difficulty with thick clouds. The third category is the multitemporal-based approaches which reconstruct the cloud-contaminated regions by fusing multi-temporal images [4-6]. The multi-temporal based methods are more intensively studied and more effective to cope with thick clouds, comparing with the other two categories of methods mentioned above.

This paper not only contributes to a summarization of the current multitemporal-based methods, but also proposes a promising idea of fusing multi-source and multi-temporal images for cloud removal, which aim at promoting the utilization of multi-source observation data and then improving the productivity and precision of the cloud removal methods.

## 2. MULTI-TEMPORAL METHODS FOR CLOUD REMOVAL

Satellite remote sensing systems with a fixed repeat cycle can easily acquire multi-temporal images in the same area. As the mobility of clouds, the cloud cover area of the multi-temporal images cannot just completely overlap, which is the data source to reconstruct missing information. The existing multitemporal-based methods are mostly based on the multi-temporal images acquired from the same sensor, with the consideration of the images from the same sensor sharing the same system characteristics, such as spatial resolution, bandwidth and spectral response function.

According to the main source of the filled information, the current multitemporal-based methods can be classified into three categories. The first category is temporal-replacement approach. As for this approach, the cloud-contaminated regions are directly replaced with the information from the reference image, followed by brightness adjustment [6-7]. That is to say, the brightness adjustment is after the replacement, and the information outside the "cloud region" in the reference image is not be used in this category of approach. The second category is called as integration-prediction approach. The information from the reference image is adjusted by using both the target and reference images before it is used to fill the missing regions [4-5]. In this category of approach, more information is used to calculate the missing data than the first category of approach. The third category is named as self-replacement with temporal guidance approach. This category of approach fills the cloud-contaminated information with the information from the remaining regions of the target image itself, guided by a reference image [8].

We conducted two groups of simulated data experiments to compare and analysis the three categories of methods mentioned above. The Poisson method [7], the weighted linear regression (WLR) method [5], and the spatio-temporal markov random fields (STMRF) method [8] are chosen as three representative methods of the three categories, respectively. Fig. 1 and Fig. 2 show the test data and results of the two experiments, and Table 1 lists the quantitative assessment values of the two experimental results. From Fig. 1, Fig. 2 and Table 1, it can be seen that, when the land cover change is small, such as in Fig.1, all the three methods can obtain satisfactory results, and the quantitative values of the WLR method is relatively better than the other two methods in this situation; while when the land cover change is obvious, such as in Fig.2, the STMRF can obtain better reconstruction result with higher degree of spectral coherence than the other two methods. The two experiments also suggest that the performance of multi-temporal method is related to the spatial resolution, the STMRF method tends to have a better performance in low resolution images.

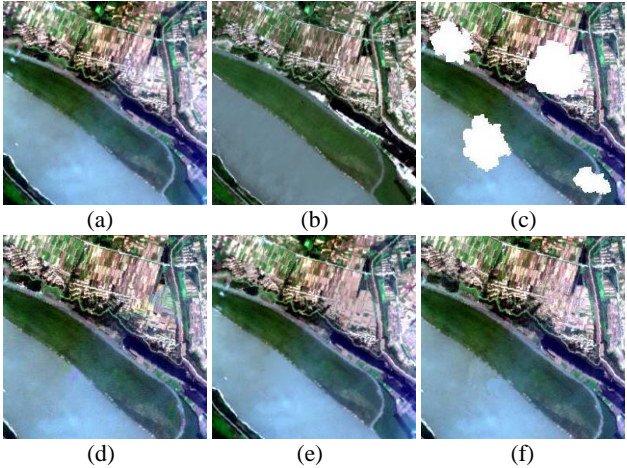

Fig. 1. Simulated data experiment 1. (a) Original GF-1 image on August 7, 2015. (b) Reference image on August 3, 2015. (c) Simulated cloud-contaminated image. Reconstruction result of (d) Poisson method. (e) WLR method. (f) STMRF method.

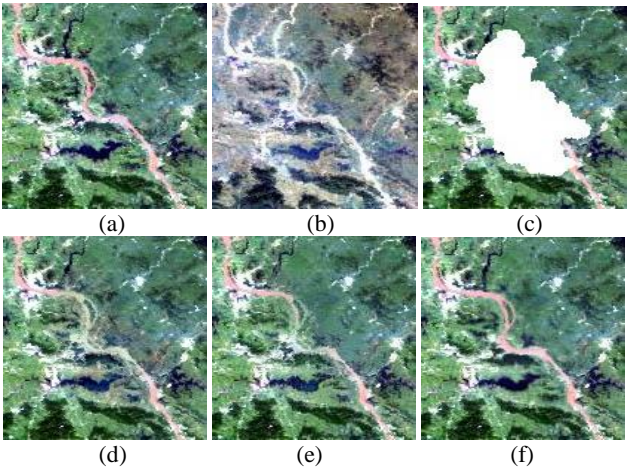

Fig. 2. Simulated data experiment 2. (a) Original MODIS image on August 2, 2010. (b) Reference image on March 11, 2010. (c) Simulated cloud-contaminated image. Reconstruction result of (d) Poisson method. (e) WLR method. (f) STMRF method.

Table 1. Quantitative assessment of the results in Fig.1 and Fig.2.

|  |  | Poisson | WLR | STMRF |
| --- | --- | --- | --- | --- |
| Fig.1 | CC | 0.8922 | **0.9228** | 0.9084 |
|  | NMSE | 0.03451 | **0.0250** | 0.0295 |
|  | UIQI | 0.8965 | **0.9184** | 0.9061 |
| Fig.2 | CC | 0.6938 | 0.7722 | **0.7941** |
|  | NMSE | 0.1257 | 0.0675 | **0.0521** |
|  | UIQI | 0.6992 | 0.7538 | **0.7815** |

## 3. MULTI-SENSOR METHODS FOR CLOUD REMOVAL

When the time interval of the multi-temporal images available is too long, the land cover may undergo significant changes. The current multitemporal-based methods generally have low accuracy in this situation. In this case, another sensor data with low spatial resolution but high temporal frequency will be useful, and the spatiotemporal fusion methods [8] can be introduced for information reconstruction. We thus propose a spatiotemporal-fusion with poisson-adjustment method for cloud removal.

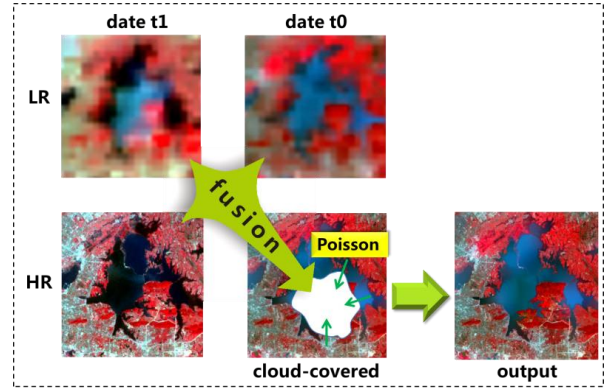

Fig. 3. The flowchart of the proposed method

The basic idea of the proposed method is as follows. As shown in Fig. 3, we refer to the cloud-covered image as high-resolution (HR) image acquired at the date $t_0$, and the auxiliary data we used are a pair of cloud-free low and high-resolution images acquired at the reference date $t_1$ and a cloud-free low-resolution image acquired near the date $t_0$. The spatiotemporal fusion algorithm [9] is used to obtain the preliminary prediction of the missing information in the cloud-covered region. Then the Poisson method [7] is followed to adjust the preliminary prediction values in accordance with the remaining regions of the cloud-contaminated image, so we get the final reconstruction results.

We undertook two simulated data experiments to test and verify the efficacy of the proposed method. In the first experiment, we simulated a cloud-covered region in a Landsat image acquired on January 13, 2005. The original image and the simulated cloud-covered image are shown in Fig. 4(e) and 4(a) respectively. The auxiliary images are a MODIS image acquired on the same data, and a pair of Landsat and MODIS images acquired on October 25, 2004, as shown in Fig. 4(b)-(d). The proposed method is compared with the three methods mentioned above. The recovery results of each method are shown in Fig. 4(f)-(i), and the detailed regions in the original image and the recovery results are shown in Fig. 4(j)-(n). It can be seen that the target and auxiliary images undergo significant changes. The three multi-temporal based methods cannot deal with this obvious change problem very well in terms of spectral coherence and spatial details as shown in Fig. 4(k)-(m). The proposed method is better able to address this issue, and the recovery result (Fig. 4(n)) is much closer to the original Landsat image (Fig. 4(j)). The effectiveness of the proposed method can also be illustrated by the quantitative assessment listed in Table 2. We can see that, for the result of the proposed method, the value of NMSE (normalized mean square error) is much lower, CC (correlation coefficient) and UIQI (universal image quality index) are much higher than other three methods.

The reconstruction ability of the proposed method in heterogeneous regions is also illustrated. In the second experiment, the study area is more spatially fragmentary and sporadic parcels are distributed around. Cloud contamination is simulated in the Landsat image on January 5, 2002, and the acquisition data of auxiliary is April 2, 2002. All original, cloud-simulated and reconstructed image along with zoomed-in subsets are shown in Fig. 5, displayed as Fig. 4. Among recovery images shown in Fig. 5(f)-(i), the WLR method obtains a relatively satisfactory result. In contrast, both Poisson method and the proposed method suffer from spectral inconsistency with the target image, and the STMRF method cannot preserve spatial distribution of features well. The unsatisfactory result of the proposed method is mainly caused by two reasons. Firstly, MODIS pixels acquired over spatially heterogeneous areas are spectrally mixed, which causes that temporal change information cannot be accurately provided in the MODIS images. Secondly, the auxiliary Landsat and MODIS data suffer from radiometric inconsistency in this experiment, as shown in Fig. 5(c) and 5(d), which also has a side-effect on prediction accuracy. Quantitative assessment in Table 2 also indicates the proposed method has a limitation for reconstruction over heterogeneous landscapes.

Table 2. Quantitative assessment of the results in Fig.4 and Fig.5.

|  |  | Poisson | WLR | STMRF | Proposed |
|---|---|---|---|---|---|
| Fig.4 | CC | 0.6662 | 0.7177 | 0.6946 | **0.8411** |
|  | NMSE | 0.0502 | 0.0434 | 0.0616 | **0.0260** |
|  | UIQI | 0.6268 | 0.7107 | 0.6909 | **0.8135** |
| Fig.5 | CC | 0.3894 | **0.6689** | 0.5861 | 0.5780 |
|  | NMSE | 0.6852 | **0.0670** | 0.0900 | 0.0938 |
|  | UIQI | 0.3404 | **0.6610** | 0.5682 | 0.4939 |

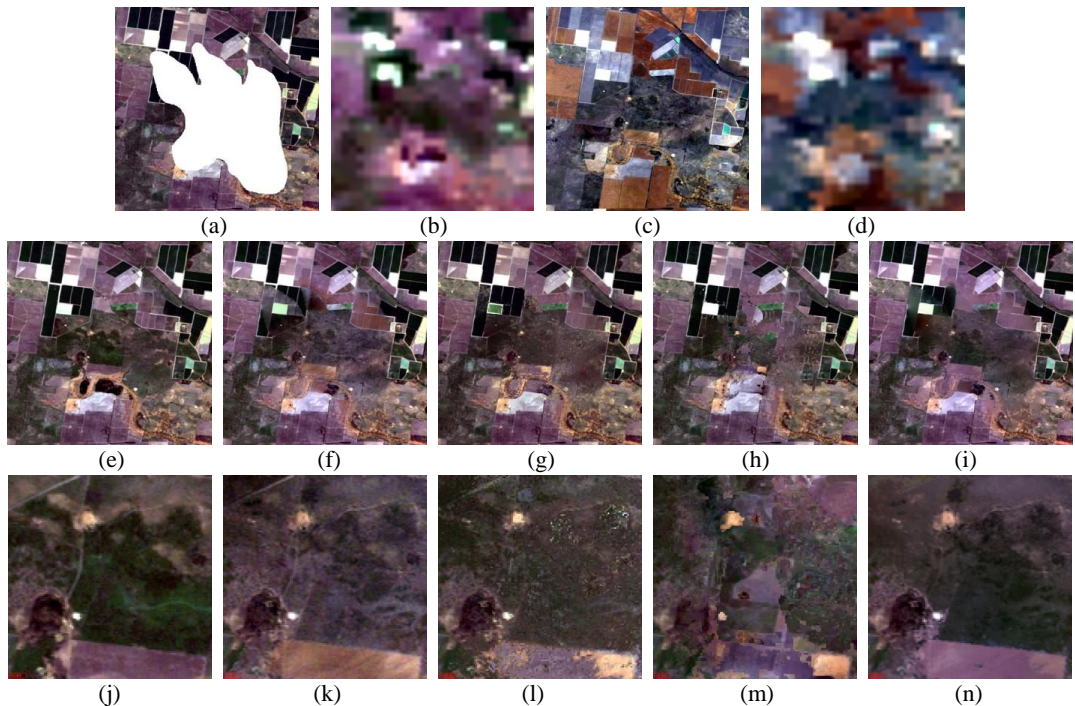

Fig. 4. Test data and results. (a) Simulated cloud-contaminated Landsat image on January 13, 2005. (b) Auxiliary MODIS image on January 13, 2005. (c)-(d) Auxiliary Landsat and MODIS images on October 25, 2004. (e) Original Landsat image on January 13, 2005. Result of (f) Poisson method. (g) WLR method. (h) STMRF method. (i) The proposed method. (j)-(n) The detail of (e)-(i).

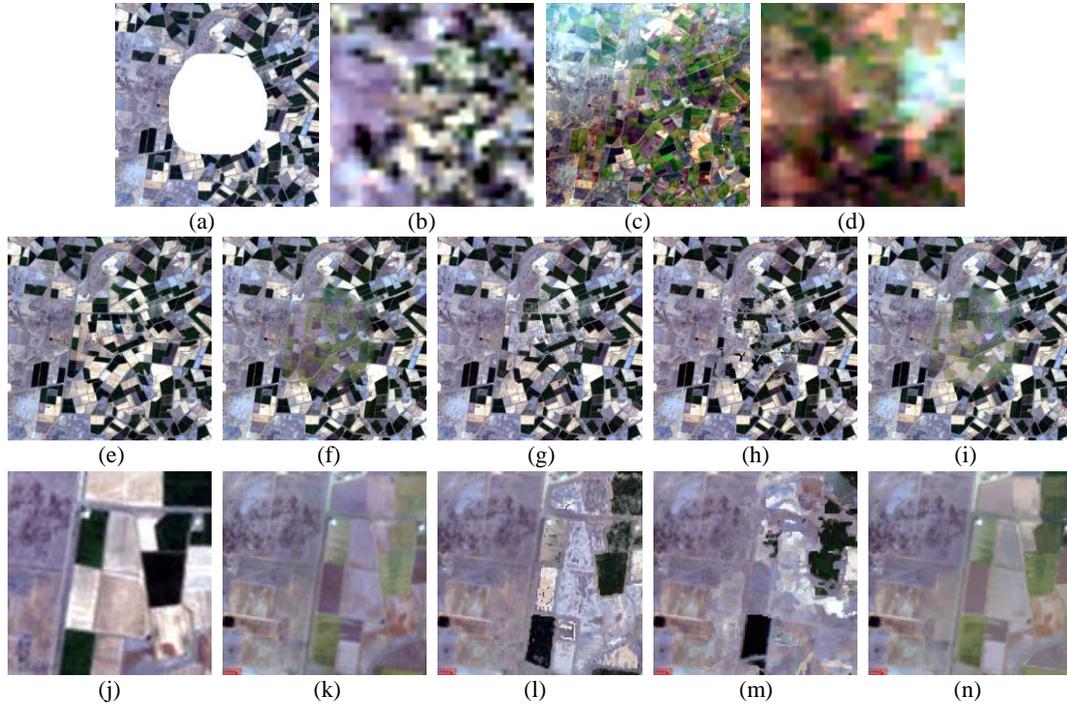

Fig. 5. Test data and results. (a) Simulated cloud-contaminated Landsat image on January 5, 2002. (b) Auxiliary MODIS image on January 5, 2002. (c)-(d) Auxiliary Landsat and MODIS images on April 2, 2002. (e) Original Landsat image on January 5, 2002. Result of (f) Poisson method. (g) WLR method. (h) STMRF method. (i) The proposed method. (j)-(n) The detail of (e)-(i).

## 4. CONCLUSION

This paper summarized the existing multitemporal-based cloud removal approaches, and classified them into three categories: the temporal-replacement method, the integration-prediction method, and the self-replacement with temporal guidance method. Moreover, a spatiotemporal-fusion with poisson-adjustment method was proposed in this paper to fuse multi-sensor and multi-temporal images for cloud removal. This proposed method introduced the spatiotemporal-fusion technique to reconstruct the missing information in the cloud-contaminated regions, following by a Poisson method to adjust the preliminary reconstruction values in accordance with the remaining regions of the cloud-contaminated image. The experiment results show that the proposed method has a potential advantage to deal with the significant changes of the multi-temporal images. However, its reconstruction accuracy degrades when used in heterogeneous regions.